\documentclass[journal]{IEEEtran}

\usepackage{cite}
\usepackage{times}
\usepackage{epsfig}
\usepackage{graphicx}
\usepackage{amsmath,amsfonts}
\usepackage{amssymb,amsopn}
\usepackage{amsthm}
\usepackage{textcomp}
\usepackage{xspace}
\usepackage{booktabs}
\usepackage{longtable}
\usepackage{multirow}
\usepackage{gensymb}
\usepackage{color,xcolor}
\usepackage[table]{xcolor}
\usepackage{epstopdf}
\usepackage{algorithm}
\usepackage{algorithmic}
\usepackage{diagbox}
\usepackage{bm}
\usepackage{url}

\newcommand{\mat}[1]{\boldsymbol{#1}}

\newcommand{\eat}[1]{}
\providecommand{\eg}{\emph{e.g.}\xspace}
\providecommand{\ie}{\emph{i.e.}\xspace}

\providecommand{\etal}{\emph{et al.}\xspace}
\newlength\savewidth\newcommand\shline{\noalign{\global\savewidth\arrayrulewidth
  \global\arrayrulewidth 1.5pt}\hline\noalign{\global\arrayrulewidth\savewidth}}
\newcommand{\longrightarrowtext}[1]{\xrightarrow{\hspace{#1}}}
\newcommand{\methodname}{{GraphMAR}\xspace}
\newcommand{\methodnamenospace}{{GraphMAR}}

\newcommand{\modelname}{{GraphMoE}\xspace}

\renewcommand{\paragraph}[1]{\noindent\textbf{#1}\hspace{2mm}}
\newcommand{\papertitle}{\methodname: Geometry-Aware Graph Learning Framework for Spatially Adaptive CT Metal Artifact Reduction}
\usepackage{hyperref}
\hypersetup{colorlinks=true,linkcolor=red,anchorcolor=blue,citecolor=blue}

\begin{document}

\title{\papertitle}

\author{Zilong~Li, Chenglong~Ma, Yiming~Lei, Yuanlin~Li, Jing~Han, Jiannan~Liu, Huidong~Xie, Junping~Zhang, Yi~Zhang, and Hongming~Shan%
\thanks{Z. Li and J. Zhang are with the Shanghai Key Lab of Intelligent Information Processing, College of Computer Science and Artificial Intelligence, Fudan University, Shanghai 200433, China. (e-mail: longzilipro@gmail.com)}%
\thanks{C. Ma, H. Xie, and H. Shan are with the Institute of Science and Technology for Brain-inspired Intelligence, Fudan University, Shanghai 200433, China. (e-mail: hmshan@fudan.edu.cn)}%
\thanks{Y. Lei is with the College of Computer Science and Technology, Qingdao University, Qingdao 266071, China.}%
\thanks{Y. Li, J. Han, and J. Liu are with the Department of Oral Maxillofacial Head and Neck Oncology, Shanghai Ninth People's Hospital, Shanghai Jiao Tong University School of Medicine, Shanghai 200011, China.}%
\thanks{Y. Zhang is with the School of Cyber Science and Engineering, Sichuan University, Sichuan 610065, China.}}

\maketitle

\begin{abstract}
Computed tomography (CT) metal artifact reduction (MAR) aims to reduce the severe streaking artifacts induced by metallic implants and other high-density objects. 
Effective MAR generally requires both accurate artifact localization and artifact removal. 
Sinogram-domain methods can exploit explicit geometric cues, such as metal traces, to identify metal-corrupted measurements, but they require access to raw projection data, which is often unavailable in clinical and practical scenarios. 
Image-domain methods are therefore more flexible and widely applicable, yet they usually lack comparable geometric guidance, limiting their ability to localize artifacts and often leading to suboptimal results.
To address this limitation, we propose \methodname, a geometry-aware learning framework for explicit artifact identification and spatially adaptive MAR in the image domain.
The key idea is to introduce graph-based geometric modeling as an image-domain analogue of sinogram metal traces.
Specifically, we first construct a geometric graph from the metal mask and derive a geometric density graph that coarsely localizes artifact-prone regions according to inter-implant geometry.
We then design \modelname, a graph-routed mixture-of-experts module that builds a polar-coordinate artifact graph in feature space and adaptively routes different experts to different spatial regions for MAR. 
By aligning the learned routing maps with the geometric density graph, \methodname provides explicit and interpretable artifact localization while enabling region-adaptive artifact reduction.
Experiments on both simulated and real-world datasets demonstrate that \methodname achieves superior MAR performance compared with existing methods. 
In addition, the proposed framework is computationally efficient and can be readily integrated with various MAR backbones. 
To the best of our knowledge, this is the first work to introduce graph-based modeling for CT MAR and to enable explicit artifact identification in the image domain, improving both restoration quality and interpretability. 
Source code will be released upon acceptance of this work.
\end{abstract}


\section{Introduction}
\begin{figure}[t]
    \centering    
    \includegraphics[width=1\linewidth]{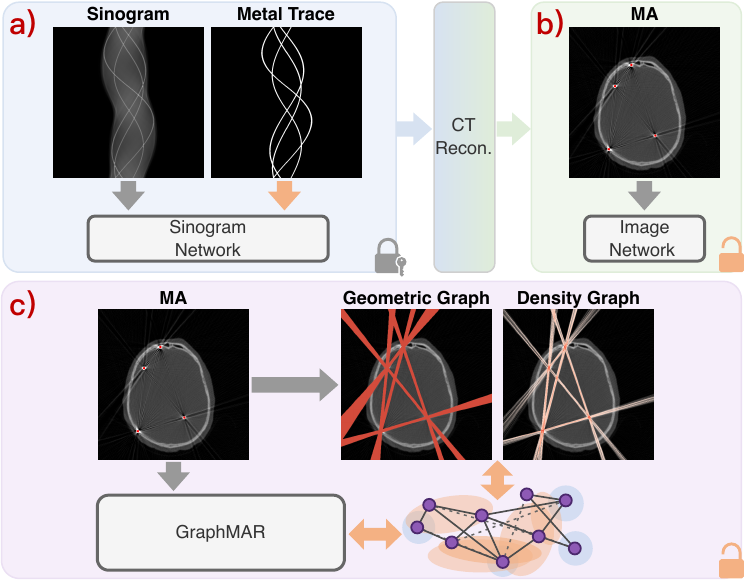}
    \caption{
    Illustration of different CT MAR paradigms. 
    \textbf{a)} Sinogram-domain methods restore corrupted sinograms using the metal trace as guidance, but may introduce secondary artifacts and require raw projection data, which are difficult to obtain in clinical scenarios.
    \textbf{b)} Conventional image-domain methods implicitly identify artifacts and restore images, often producing oversmoothed results with limited interpretability.
    \textbf{c)} \methodname does not rely on raw projection data and constructs image-domain geometric and density graphs to guide explicit artifact identification and adaptive MAR. 
    }
    
    \label{fig:motivation} 
\end{figure}

Computed tomography (CT) is widely used for clinical diagnosis and screening. However, high-density implants, such as dental fillings and metal prostheses, can corrupt the CT raw data, i.e., the sinogram, producing severe streaking artifacts that propagate across the reconstructed image and obscure anatomical structures.
To mitigate this problem, metal artifact reduction (MAR) algorithms~\cite{MAR_40years, aapm_challenge} have been developed in both the sinogram and image domains to suppress artifacts and recover clinically relevant details. Since metal-induced artifacts in reconstructed images are highly non-local and difficult to localize, most state-of-the-art methods perform artifact reduction in the sinogram domain~\cite{prior,dudonet,dudonet++,indudo}. In this domain, metal-corrupted measurements can be explicitly identified using the metal trace, a reliable geometric prior derived from CT imaging geometry, as illustrated in Fig.~\ref{fig:motivation}.

While sinogram-domain methods benefit from explicit geometric cues such as metal traces, they can introduce secondary artifacts after reconstruction~\cite{dudonet, quadnet}. More importantly, once transformed into the image domain, both original metal artifacts and secondary artifacts become difficult to localize. As a result, explicit artifact guidance remains largely unavailable in the image domain, posing a common challenge for both image-only and dual-domain MAR methods.  
The central difficulty is distinguishing metal-induced artifacts from true anatomical structures in reconstructed CT images. Without a clear geometric guide analogous to the metal trace, image-domain networks face three key limitations: (\textbf{i}) they must implicitly separate complex artifacts from anatomical content, which is challenging and unreliable; (\textbf{ii}) they tend to process all regions uniformly, often oversmoothing clinically important details while insufficiently suppressing severe artifacts; and (\textbf{iii}) they largely overlook the intrinsic geometric patterns underlying metal artifacts.

To address these challenges, we ask: \emph{Can geometric guidance be introduced into the image domain to enable explicit artifact identification and adaptive MAR?} Our key insight is that severe metal artifacts are not randomly distributed in reconstructed CT images; instead, they exhibit stable geometric patterns governed by the CT reconstruction process. As illustrated in Fig.~\ref{fig:motivation}(Bottom), strong streak artifacts tend to emerge along paths connecting metal implants. By representing these inter-implant relationships as a graph, we can coarsely localize artifact-prone regions and provide explicit geometric guidance for image-domain MAR.

Motivated by this observation, we propose \methodname, a novel framework that injects geometric knowledge into existing MAR models for explicit artifact identification and spatially adaptive artifact reduction. \methodname operates in the image-domain feature space and consists of two main components. 
First, we construct a geometric graph from the metal mask and derive a geometric density graph to coarsely localize artifact-prone regions based on inter-implant geometry. This density graph serves as an image-domain counterpart to the metal trace in the sinogram domain, providing explicit geometric guidance without requiring access to raw projection data.
Second, we design \modelname, a graph-routed mixture-of-experts module that constructs a polar-coordinate artifact graph in feature space. Based on this graph, a graph router adaptively dispatches a set of lightweight experts, each specializing in regions with different artifact severities. The router is supervised by the geometric density graph, encouraging the learned artifact attention to faithfully reflect the underlying geometric prior.
As a result, \methodname is lightweight, plug-and-play, and consistently improves a variety of MAR backbones, including both image-domain and dual-domain models. Moreover, the artifact attention map produced by \modelname{} provides an interpretable visual cue for artifact localization. This interpretability further allows clinicians to selectively control restoration strength, helping reduce the risk of hallucinated structures and secondary artifacts.

\begin{figure*}[t]
    \centering    
    \includegraphics[width=1\linewidth]{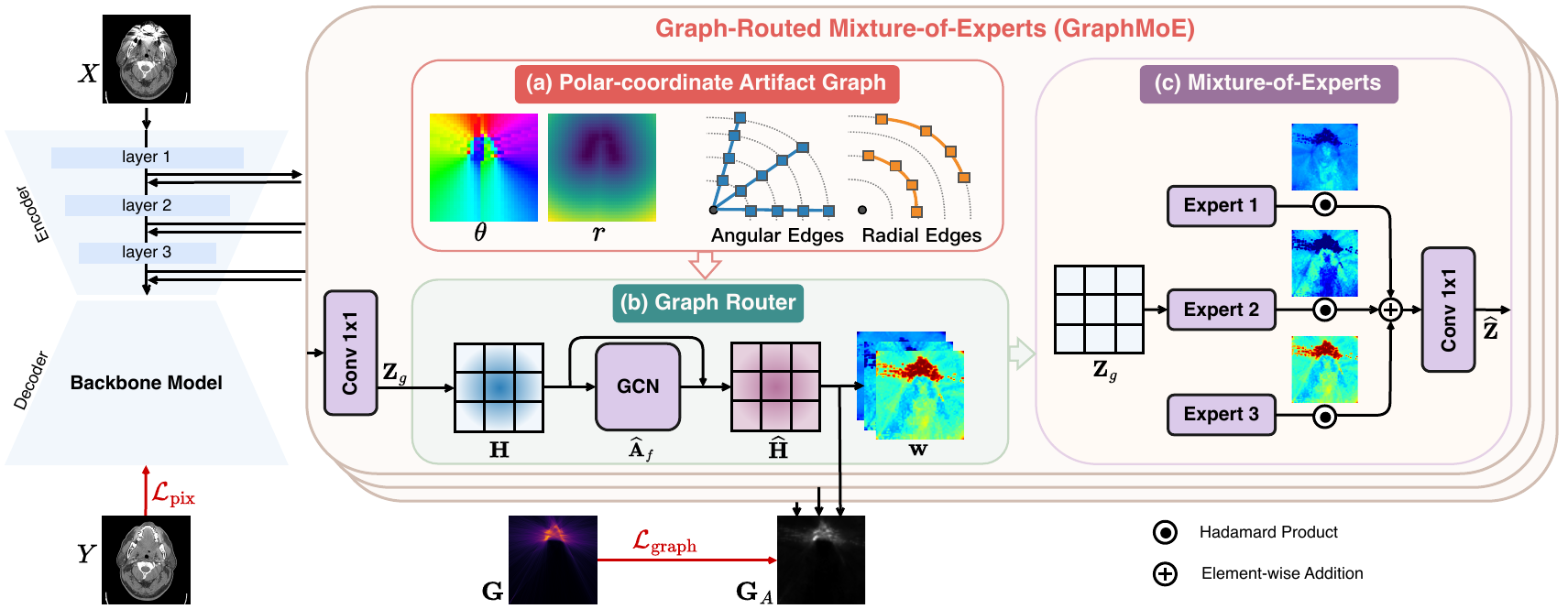}
    \caption{Overview of \methodname, which contains multiple \modelname modules integrated into the backbone model. First, a polar-coordinate artifact graph is constructed to capture the spatial geometry of artifacts, which guides the graph router to generate the routing maps. Second, a Mixture-of-Experts module uses these maps to adaptively fuse expert features for spatially adaptive restoration. Finally, the output feature $\widehat{\mat{Z}}$ is added to the backbone feature map via a residual connection. We supervise the graph router via the precomputed geometric density graph $\mat{G}$. }

    \label{fig:framework} 
\end{figure*}

Our contributions are summarized as follows.
\begin{enumerate}
    \item We propose \methodname, a geometry-aware graph learning framework for spatially adaptive CT MAR that enables explicit artifact identification and spatially adaptive artifact reduction in the image domain. To the best of our knowledge, this is the first work to introduce graph-based modeling for MAR.
    \item We introduce a geometric graph that captures inter-implant relationships and robustly highlights artifact-prone regions, serving as an image-domain geometric prior analogous to the metal trace in the sinogram domain.
    \item We design \modelname, a plug-in graph-routed mixture-of-experts module that constructs a polar-coordinate artifact graph in feature space and adaptively routes lightweight experts for region-specific MAR.
    \item Extensive experiments on simulated and real-world datasets demonstrate that \methodname consistently improves MAR performance while producing interpretable artifact attention maps, enhancing the clinical reliability of image-domain restoration.
\end{enumerate}

\section{Related Work}

\subsection{Metal Artifact Reduction}
Metal artifacts in CT images originate from complex physical phenomena such as beam hardening and photon starvation~\cite{correctionMAR1, correctionMAR2, correctionMAR3, correctionMAR4}. Initially corrupting part of the data in the sinogram, these artifacts propagate throughout the entire image after reconstruction. 
Traditional methods like Linear Interpolation (LI)~\cite{LI}, NMAR~\cite{NMAR}, and CNNMAR~\cite{cnnmar}, treat MAR as a sinogram interpolation problem, but often introduce severe secondary artifacts due to imprecise interpolation.
In contrast, image-domain deep learning methods directly model the implicit relationship between artifact images and corresponding ground truths, effectively enhancing the image quality, but may also suffer from oversmoothing, and fail when the artifact is complicated~\cite{cnnmar, cyclegan, adn, hongacdnet, hongdicdnet}. To address these issues, dual-domain methods simultaneously address artifacts in both sinogram and image domains, leading to state-of-the-art results~\cite{dudonet, indudo, dudonet++, dannet, prior, su2024f2iflow}.

Although leading deep learning methods achieve promising visual quality, they generally lack explicit artifact localization. This limitation can introduce secondary artifacts or oversmooth critical anatomical details, reducing clinical reliability~\cite{quadnet, risemar}. Conventional methods such as Frequency-split NMAR address this issue through a divide-and-conquer strategy that separately processes regions near and far from metal implants, thereby improving reliability via explicit artifact modeling~\cite{FSNMAR, fsnmargood1, fsnmargood2}. 
In contrast, \methodname brings explicit geometric guidance into the image domain, enabling artifact-aware localization and spatially adaptive restoration within a learning-based MAR framework.

\subsection{Graph for Image Modeling}
Geometric structures are prevalent in images, including topological relationships and connections among image features. Such structures can be effectively represented by graphs, where pixels or patches are treated as nodes and their similarities or relationships are modeled as edges~\cite{vig, vihgnn}. Graph-based methods have shown strong potential in low-level vision tasks, including image restoration~\cite{lowlevel_eth_graph, lowlevel_graph}, super-resolution~\cite{lowlevel_graph_sr}, and image inpainting~\cite{inpainting_scenegraph, inpainting_graph}.
Graph neural networks (GNNs) provide a powerful framework for processing graph-structured data by aggregating messages along edges and updating node embeddings~\cite{GCN, GAT, vihgnn}. Compared with conventional vision models, GNNs can capture long-range dependencies more effectively than CNNs~\cite{cnn}, while avoiding the high computational cost of dense attention mechanisms~\cite{attention, vit}.

In MAR, metal artifacts often propagate along specific directions and exhibit distinct geometric patterns governed by CT imaging geometry. These properties make graph modeling particularly suitable for artifact localization and reduction. Motivated by this observation, we design a geometric graph for explicit artifact localization and a polar-coordinate artifact graph for spatially adaptive MAR.

\section{Method}
\subsection{Overview of \methodname}

Fig.~\ref{fig:framework} presents an overview of \methodname, which introduces explicit artifact identification and spatially adaptive MAR. 
First, a geometric density graph $\mat{G}$ is computed from the metal mask via classical image processing techniques. The density graph provides a coarse and stable prior that localizes artifact-affected regions. 
Second, we design graph-routed mixture-of-experts, \modelname. A polar-coordinate artifact graph is first constructed in feature space to capture artifact geometry. A graph router then processes it to produce a routing map. A mixture-of-experts uses the routing map to adaptively restore different regions. 
To supervise the routing with the geometric prior, we aggregate the routing maps into an artifact attention map $\mat{G}_A$ and align it with the geometric density graph via a geometric alignment loss. 
Finally, \modelname produces an enhanced feature $\widehat{\mat{Z}}$ that is added back to the backbone via a residual connection, while $\mat{G}_A$ provides an interpretable visual cue for artifact-affected regions.
Since \methodname operates entirely in the image-domain feature space, it readily adapts to both image- and dual-domain backbones.

We use $X$ to denote the input CT image, $Y$ its artifact-free ground truth, and $\widehat{Y}$ the network prediction. Next, we detail each component in \methodname.

\subsection{Geometric Graph}
\label{sec:geometric_graph}

In the sinogram domain, metal artifacts can be localized through metal traces or metal-mask projections~\cite{dudonet,cnnmar,dudonet++,quadnet}, both of which derive from CT geometry. This geometric prior allows sinogram-domain methods to focus on artifact-affected regions while leaving other areas unchanged. No comparable prior is available in the image domain.
Since CT reconstruction can be viewed as the superposition of a series of line integrals, severe artifacts tend to concentrate along paths between metal implants. Based on this observation, we construct a geometric graph from the metal mask using classical image processing techniques to coarsely localize artifact-prone regions.
Fig.~\ref{fig:artifact_graph} illustrates the construction, which involves three steps: \emph{node extraction}, constructing a \emph{geometric graph}, and deriving a \emph{geometric density graph}. Note that the density graph $\mat{G}$ is derived from the geometric graph to better fit MAR networks. We detail the three steps as follows:

\subsubsection{Node extraction}
First, we threshold the CT image at 2800 Hounsfield Units (HU) to obtain a binary metal mask $\mat{M}$~\cite{dudonet, quadnet}. Connected-component labeling then partitions $\mat{M}$ into $N$ separate metal regions, one per implant. To reduce the computational cost, we retain only the boundary pixels of each region, forming pixel sets $P=\{P_1,\ldots,P_N\}$, where $P_i=\{p_{i1},\ldots,p_{in_i}\}$ contains the boundary pixels of the $i$-th implant.

\subsubsection{Geometric graph}
For any two implants $i$ and $j$, we pair boundary pixels $p_{ik} \in P_i$ and $p_{jl} \in P_j$ to create an edge. Let $\mathcal{S}=\{(p_{ik}, p_{jl}) \mid p_{ik}\in P_i,\ p_{jl}\in P_j,\ i\neq j\}$ denote the set of all inter-implant boundary-point pairs. The resulting geometric graph comprises all boundary pixels as nodes and all inter-implant pairs as edges.

\subsubsection{Geometric density graph}
\begin{figure}[t]
    \centering    
    \includegraphics[width=1.0\linewidth]{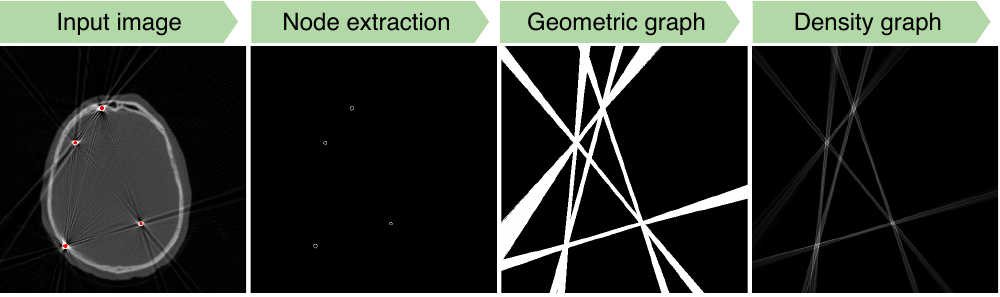}
    \caption{The procedure of constructing the geometric graph and geometric density graph. Boundary pixels of each implant are extracted as graph nodes (\emph{node extraction}); straight lines between boundary pixels from different implants form graph edges (\emph{geometric graph}); the per-pixel traversal count is then min--max normalized to $[0,1]$, yielding a continuous density graph $\mat{G}$ where brighter pixels indicate regions more likely affected by artifacts (\emph{geometric density graph}).}
    \label{fig:artifact_graph} 
\end{figure}

Since most MAR backbones operate on image grids rather than graph-structured data, we convert the geometric graph into a spatial density graph. Starting from a zero map, we trace the line segment of each pair in $\mathcal{S}$ and add $1$ to every pixel it traverses. The accumulated map is then min--max normalized to $[0,1]$ to yield the geometric density graph $\mat{G}$. As illustrated in Fig.~\ref{fig:artifact_graph}, $\mat{G}$ robustly covers the regions with severe artifacts.

\subsection{Graph-Routed Mixture-of-Experts (\modelname)}
Although $\mat{G}$ robustly highlights the most severely affected regions, it is coarse and image-agnostic, as it derives only from the metal mask. Therefore, we further introduce graph-routed mixture-of-experts (\modelname{}), which adaptively learns artifact patterns in feature space and takes $\mat{G}$ as guidance.

\modelname{} consists of two components that play different roles: (1) \emph{graph router} constructs a polar-coordinate artifact graph to capture spatial artifact geometry in the feature map and produces a routing map; and (2) the \emph{mixture-of-experts} module uses the routing map to adaptively restore different regions.

\subsubsection{Polar-coordinate artifact graph}
\label{sec:artifact_graph}

Due to the long-range propagation of metal artifacts, CNN struggles to capture their long-range structures, while transformer is computationally prohibitive for high-resolution images~\cite{clma_proct, quadnet}. To effectively model these non-local dependencies, we design a sparse artifact graph in the polar-coordinate space tailored to the artifact geometry. 
 As the spatial distribution of metal artifacts is closely tied to the positions of metal implants, two characteristic patterns can be observed. First, pixels along the same X-ray projection direction from an implant tend to exhibit similar streak artifacts. Second, pixels at a similar distance from the metal share comparable artifact intensity. 
Inspired by this, we compute local polar coordinates for each feature point relative to its nearest metal pixel.

\begin{figure}[t]
    \centering
    \includegraphics[width=1\linewidth]{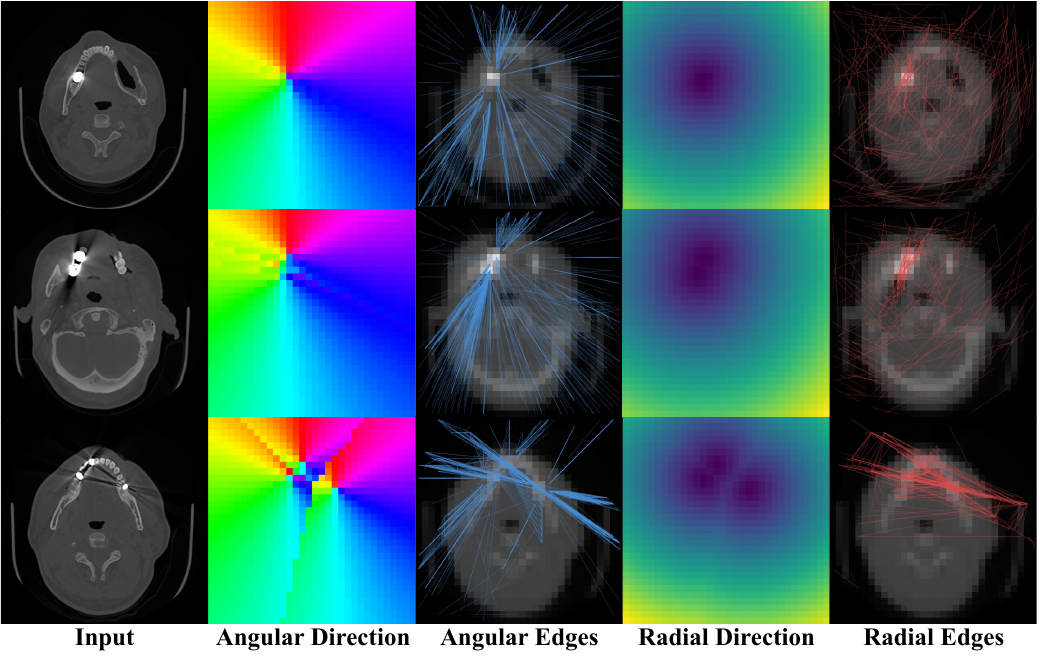}
    \vspace{-8pt}
    \caption{Illustration of the polar coordinate and the two types of edges in the artifact graph. Each feature point is assigned a radial distance $r$ and an angular direction $\theta$ relative to its nearest metal pixel. Angular edges (blue) connect nodes sharing similar streak directions, while radial edges (red) connect nodes at similar distances from the metal.}
    \label{fig:artifact_graph_construction}
\end{figure}

Algorithm~\ref{alg:graph_construction} details the full procedure.
First, the metal mask $\mat{M}$ and geometric density graph $\mat{G}$ are resized to the feature resolution to obtain the metal set $\mat{M}_f$  and density graph $\mat{G}_f$ at feature scale. 
Second, for each feature point $v_i$ in the feature map, we find its nearest metal pixel $p_i^*=\operatorname*{arg\,min}_{p\in\mat{M}_f}\|v_i-p\|_2$ and compute a relative spatial vector $\mat{d}_i = v_i - p_i^*$. The polar coordinates of the feature point $v_i$ relative to the metal implant $p_i^*$ are then defined as:
\begin{align}
    \theta_i = \operatorname{atan2}(d_{i,y},\, d_{i,x}), \qquad
    r_i = \| \mat{d}_i \|_2,  
    \label{eq:polar}
\end{align}
where $d_{i,x}$ and $d_{i,y}$ are the horizontal and vertical components of $\mat{d}_i$. 
Fig.~\ref{fig:artifact_graph_construction} illustrates the angular direction $\theta$ and radial distance $r$ of each pixel. Based on the polar coordinates, we construct two types of edges to capture the artifact geometry: 
\begin{itemize}
    \item \textbf{Angular edges} connect pixels along the same streak direction. For each node, we use a Gaussian kernel $w^{\theta}_{ij} = \exp(-d_\theta(i,j)^2 / {2\sigma^2})$ to measure how close two nodes are in angular direction, where $d_\theta(i,j)$ is the circular distance between $\theta_i$ and $\theta_j$. 
    \item \textbf{Radial edges} connect pixels at the same distance from the metal. Similarly, we use $w^{r}_{ij} = \exp(-(r_i - r_j)^2 / 2\sigma^2)$ to measure how close two nodes are in radial distance.
\end{itemize}
To reduce computational cost and ensure graph sparsity, we construct edges by connecting each node to its most similar peers, specifically, 12 other nodes for angular similarity and 4 nodes for radial similarity. We empirically set the scaling factors to $\sigma = 2.0$ to get the initial artifact graph $\mat{A}_f$.

Third, to further concentrate the polar-coordinate graph on artifact-affected regions, each edge is reweighted by the geometric density graph, i.e., $\sqrt{\mat{G}_f(v_i)\cdot\mat{G}_f(v_j)}$, when multiple implants are present ($N \geq 2$). 
Finally, we symmetrize and normalize the adjacency matrix to get the final artifact graph $\widehat{\mat{A}_f}$~\cite{GCN}. As a result, the artifact graph encodes the geometric structure of artifacts, serving as a robust spatial prior for the subsequent graph routing.

\begin{algorithm}[t]
\small
\caption{The procedure of constructing the polar-coordinate artifact graph.}
\label{alg:graph_construction}
\begin{algorithmic}[1]
\renewcommand{\algorithmiccomment}[1]{\hfill{\color{gray}\(\triangleright\) #1}}
\REQUIRE $\mat{M}$, $\mat{G}$ at image scale; feature size $H_f{\times}W_f$
\ENSURE $\mat{G}_f$, $\widehat{\mat{A}}_f$
\STATE $\mat{M}_f \gets \mathrm{Resize}_{\text{NN}}(\mat{M},\,H_f{\times}W_f)$ \COMMENT{Stage 1: resample to feature scale}
\STATE $\mat{G}_f \gets \mathrm{Resize}_{\text{bilinear}}(\mat{G},\,H_f{\times}W_f)$
\STATE $\mat{M}_f \gets \{p \mid \mat{M}_f(p)=1\}$
\STATE $\mat{A}_f \gets \mat{0}$ \COMMENT{Stage 2: polar-edge construction}
\FOR{each feature node $v_i$}
    \STATE $p_i^{*} \gets \arg\min_{p\in\mat{M}_f}\|v_i-p\|_2$
    \STATE $(r_i,\theta_i) \gets$ polar coordinates of $v_i$ relative to $p_i^{*}$ \COMMENT{Eq.~\eqref{eq:polar}}
    \STATE $\mathcal{N}^\theta_i \gets$ top-$12$ neighbors by $w^\theta_{ij}$ \COMMENT{angular edges}
    \STATE $\mathcal{N}^r_i \gets$ top-$4$ neighbors by $w^r_{ij}$ \COMMENT{radial edges}
    \STATE $\mat{A}_f[i,j] \mathrel{+}= w^\theta_{ij}$ for $j\in\mathcal{N}^\theta_i$;\; $\mat{A}_f[i,j] \mathrel{+}= w^r_{ij}$ for $j\in\mathcal{N}^r_i$
\ENDFOR
\IF{$N \geq 2$}
    \FOR{each nonzero edge $(i,j)$ in $\mat{A}_f$}
        \STATE $\mat{A}_f[i,j] \gets \mat{A}_f[i,j]\cdot\sqrt{\mat{G}_f(v_i)\,\mat{G}_f(v_j)}$
    \ENDFOR
\ENDIF
\STATE $\mat{A}_f \gets \mat{A}_f + \mat{A}_f^\top$ \COMMENT{Stage 3: normalization}
\STATE $\widehat{\mat{A}}_f \gets \mat{D}^{-1/2}\,\mat{A}_f\,\mat{D}^{-1/2}$ \COMMENT{$\mat{D}$ is the degree matrix}
\STATE \textbf{return} $\mat{G}_f,\,\widehat{\mat{A}}_f$
\end{algorithmic}
\end{algorithm}

\subsubsection{Graph router}
\label{sec:graph_feat}
\label{sec:artifact_attention}

Based on the precomputed artifact graph, we design a graph router to produce a spatially adaptive routing map for the subsequent mixture-of-experts module. The router jointly considers the geometric structure of the artifact graph and the visual features from the backbone. 
First, we use a $1{\times}1$ convolution to reduce the channels of the backbone feature $\mat{Z}\in\mathbb{R}^{C\times H\times W}$ to $\mat{Z}_g\in\mathbb{R}^{C_g\times H\times W}$, where $C_g{=}\min(C,128)$. To embed geometric context, we encode the polar coordinates using a two-layer MLP for the radial distance $r$ and a sinusoidal embedding for the streak direction $\theta$, and add them to $\mat{Z}_g$ to yield the feature embeddings $\mat{H}$.

\begin{figure}[t]
    \centering
    \includegraphics[width=0.85\linewidth]{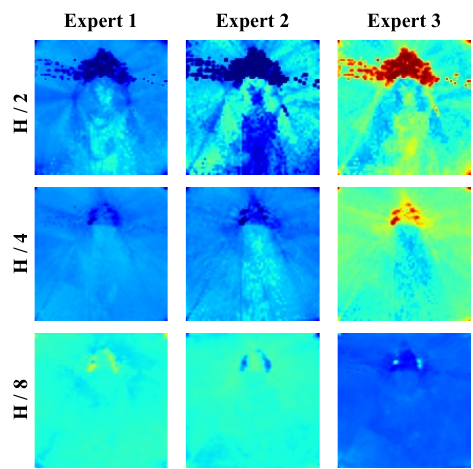}
    \caption{Visualization of the routing map of \methodname, which contains three \modelname at different scales in the backbone model.}
    \label{fig:graphmoe}
\end{figure}

Given the node embeddings $\mat{H}$ and the artifact graph $\widehat{\mat{A}}_f$, a single GCN layer performs message passing through the feature map, which can be written as:
\begin{align}
    \widehat{\mat{H}} = \operatorname{ReLU}(\operatorname{GCN}(\widehat{\mat{A}}_f,\mat{H}))+\mat{H}.
    \label{eq:gcn}
\end{align}
Finally, a zero-initialized $1{\times}1$ convolution on $\widehat{\mat{H}}$ followed by pixel-wise softmax produces the routing map $\mat{w}\in\mathbb{R}^{K\times H\times W}$ with $K{=}3$.
Fig.~\ref{fig:graphmoe} illustrates the routing maps produced by the graph router, where each routing map highlights a distinct spatial region, enabling the subsequent experts to specialize in different artifact patterns.

\subsubsection{Mixture-of-experts module}
\label{sec:expert_routing}
Guided by the predicted routing map $\mat{w}$, a lightweight mixture-of-experts module performs spatially adaptive restoration. 
First, the feature map $\mat{Z}_g$ is processed by $K$ independent experts to generate expert-specific features $\mat{U}_k = \psi_k(\mat{Z}_g)$, where $\psi_k$ denotes the $3{\times}3$ Conv--BN--ReLU block of the $k$-th expert. Next, these features are dynamically fused using the routing map $\mat{w}$: 
\begin{align}
    \widehat{\mat{Z}} = \sum_{k=1}^K \mat{w}_k \odot \mat{U}_k,
    \label{eq:moe_fusion}
\end{align}
Finally, the fused feature $\widehat{\mat{Z}}$ is projected back to $C$ channels via a $1{\times}1$ convolution and added to the original backbone feature $\mat{Z}$.

\subsection{Geometric Alignment Loss}
\label{sec:alignment}
To explicitly supervise the graph router, we map the graph embeddings $\widehat{\mat{H}}$ to a single-channel density map at each scale. These multi-scale maps are interpolated to a resolution of ($H/4, W/4$) to reduce computational cost during training. They are then concatenated and fused via a $1{\times}1$ convolution to produce a global artifact attention map $\mat{G}_A$. We compute the mean squared error between $\mat{G}_A$ and the precomputed geometric density graph $\mat{G}$, which is also interpolated to the same resolution:
\begin{align}
    \mathcal{L}_{\mathrm{graph}} = \big\|\operatorname{Norm}(\mat{G}_A) - \operatorname{Norm}(\mat{G})\big\|_2^2,
    \label{eq:mse_loss}
\end{align}
where $\operatorname{Norm}(\cdot)$ denotes min-max normalization, and this loss is omitted when only a single implant is present (\ie, $N=1$).

\subsection{Objective}
\label{sec:training}
\methodname{} is trained end-to-end with the backbone. We insert \modelname{} at the start of each encoder stage, e.g.\ at resolutions $(H_0/2,\, W_0/2)$ and $(H_0/4,\, W_0/4)$. For dual-domain networks, it is inserted only in the image-domain branch. The overall training objective combines pixel-wise reconstruction and geometric alignment:
\begin{align} 
    \mathcal{L} = \|\widehat{Y}-Y\|_1 + \lambda \mathcal{L}_{\mathrm{graph}},
    \label{eq:total_loss}
\end{align}
where $\widehat{Y}$ is the reconstructed output and we set $\lambda {=}0.1$ empirically. 

\section{Results}
\label{sec:result}

\subsection{Experimental Setup}

\subsubsection{Datasets}
We train and evaluate each method on two simulated datasets and further evaluate them on real-world a dataset.

The DDMAR dataset is generated from the DeepLesion dataset, which is widely used in previous studies~\cite{dudonet, dudonet++, quadnet}. We generate the data following the procedures outlined in QuadNet~\cite{dudonet++, quadnet}, with a training set consisting of 36,000 cases derived from 400 CT images, each containing 90 metal shapes, and a separate testing set of 2,000 cases from another 200 images, each containing an additional 10 metal shapes. The shapes of CT images and sinograms are $512 \times 512$ and $640 \times 640$, respectively.

We evaluate each method in clinically relevant dental scenarios using the private real-world dental CT dataset collected by Ma \etal \cite{risemar}, which was obtained from maxillofacial CT scanners at Shanghai Ninth People's Hospital.
The Dental dataset includes 1,125 metal artifact-affected and 2,233 unpaired artifact-free maxillofacial CT images across 104 de-identified cases.
For training and quantitative evaluation, we follow Ma \etal to create 7,200 training and 800 test image pairs from 45 clean images, utilizing metal masks randomly applied to the teeth.
In the Dental dataset, raw sinograms are not available; when sinogram-domain or dual-domain methods are evaluated, their sinograms and metal traces are generated from the reconstructed CT images through forward projection. As such a process introduces inconsistencies, we only report image-domain methods on this dataset.

In addition to the simulated evaluation, we qualitatively assess all methods on the 1,125 real-world metal-affected images from the same hospital. Since these clinical images do not have artifact-free ground truths, we provide visual comparisons to evaluate the practical effectiveness of each method.

\begin{table*}[!htb]
    \caption{Quantitative evaluation in the form of [PSNR (dB) / SSIM (\%)] for state-of-the-art methods on the DDMAR dataset. Methods are grouped into non-dual-domain and dual-domain settings, and the best result in each group is marked in \textbf{bold}.}
    \label{tab:deeple}
    \centering
    {
    \begin{tabular}{lcccccc}
    \shline
    Methods & Small Metals & \multicolumn{3}{c}{$\longrightarrowtext{3cm}$} & Large Metals &  Average \\
    \hline
    \multicolumn{7}{l}{\emph{single-domain methods}} \\
    MA
    & 32.52 / 90.26 & 31.50 / 89.23 & 30.96 / 88.43 & 27.24 / 84.17 & 23.67 / 78.54 & 29.18 / 86.13 \\
    LI~\cite{LI}
    & 35.91 / 94.58 & 35.13 / 94.13 & 33.96 / 93.72 & 30.50 / 91.99 & 28.51 / 89.95 & 32.80 / 92.88 \\
    NMAR~\cite{NMAR}
    & 36.40 / 94.71 & 35.48 / 93.68 & 34.76 / 93.85 & 31.76 / 92.71 & 29.92 / 91.19 & 33.66 / 93.23 \\
    FBPConvNet~\cite{fbpconvnet}
    & 44.43 / 98.00 & 43.82 / 97.96 & 43.24 / 97.90 & 37.58 / 97.26 & 38.59 / 96.64 & 41.53 / 97.55 \\
    ACDNet~\cite{hongacdnet}
    & 44.86 / 98.03 & 44.47 / 97.99 & 43.76 / 97.91 & 39.42 / 97.33 & 39.99 / 96.81 & 42.50 / 97.61 \\
    FredNet~\cite{gloredi}
    & 44.99 / 98.06 & 44.90 / 98.04 & 43.32 / 97.92 & 38.94 / 97.35 & 40.86 / 96.91 & 42.60 / 97.65 \\
    ProCT~\cite{clma_proct}
    & 44.46 / 97.91 & 44.23 / 97.87 & 42.62 / 97.74 & 39.53 / 97.23 & 40.62 / 96.77 & 42.29 / 97.50 \\
    \cellcolor{cyan!10}\methodnamenospace\textsubscript{ProCT}
    & \cellcolor{cyan!10}\textbf{45.88} / \textbf{98.20} & \cellcolor{cyan!10}\textbf{45.56} / \textbf{98.17} & \cellcolor{cyan!10}\textbf{44.05} / \textbf{98.04} & \cellcolor{cyan!10}\textbf{39.82} / \textbf{97.58} & \cellcolor{cyan!10}\textbf{41.63} / \textbf{97.27} & \cellcolor{cyan!10}\textbf{43.39} / \textbf{97.85} \\
    \hline
    \multicolumn{7}{l}{\emph{Dual-domain methods}} \\
    DuDoNet++~\cite{dudonet++}
    & 46.49 / 98.33 & 46.44 / 98.31 & 45.99 / 98.27 & 44.84 / 98.06 & 43.59 / 97.74 & 45.47 / 98.14 \\
    QuadNet~\cite{quadnet}
    & 46.48 / 98.35 & 46.25 / 98.32 & 45.58 / 98.25 & 43.74 / 97.99 & 42.56 / 97.56 & 44.92 / 98.09 \\
    \cellcolor{cyan!10}\methodnamenospace\textsubscript{DuDoNet++}
    & \cellcolor{cyan!10}\textbf{46.90} / \textbf{98.40} & \cellcolor{cyan!10}\textbf{46.92} / \cellcolor{cyan!10}\textbf{98.38} & \cellcolor{cyan!10}\textbf{46.36} / \textbf{98.33} & \cellcolor{cyan!10}\textbf{45.12} / \textbf{98.11} & \cellcolor{cyan!10}\textbf{43.78} / \textbf{97.77} & \cellcolor{cyan!10}\textbf{45.82} / \textbf{98.20} \\
    \shline
    \end{tabular}}
    \vspace{-4pt}
\end{table*}

\begin{table*}[!htb]
    \caption{Quantitative evaluation in the form of [PSNR (dB) / SSIM (\%)] on the Dental dataset. The best results are marked in \textbf{bold}.}
    \label{tab:teeth}
    \renewcommand{\arraystretch}{1.0}
    \centering
    {
    \begin{tabular}{lcccccc}
    \shline
    Methods & Small Metals & \multicolumn{3}{c}{$\longrightarrowtext{3cm}$} & Large Metals &  Average \\
    \hline
    MA
    & 33.71 / 97.28 & 28.97 / 94.87 & 26.75 / 93.25 & 24.15 / 90.81 & 21.10 / 86.92 & 26.94 / 92.63 \\
    LI~\cite{LI}
    & 33.46 / 94.23 & 30.69 / 91.78 & 29.14 / 90.29 & 27.69 / 88.42 & 25.76 / 86.12 & 29.35 / 90.17 \\
    NMAR~\cite{NMAR}
    & 34.88 / 93.51 & 32.12 / 91.66 & 30.46 / 90.36 & 28.81 / 88.67 & 26.45 / 86.34 & 30.55 / 90.11 \\
    FBPConvNet~\cite{fbpconvnet}
    & 40.54 / 98.85 & 38.56 / 98.37 & 37.38 / 98.07 & 36.10 / 97.63 & 34.16 / 96.88 & 37.35 / 97.96 \\
    FredNet~\cite{gloredi}
    & 44.26 / 99.03 & 41.41 / 98.53 & 39.85 / 98.23 & 38.04 / 97.77 & 35.50 / 97.03 & 39.81 / 98.12 \\
    RISE-MAR~\cite{risemar} & 41.88 / 98.98 & 39.29 / 98.44 & 37.89 / 98.09 & 36.38 / 97.58 & 34.32 / 96.80 & 37.95 / 97.98 \\
    
    ProCT~\cite{clma_proct}
    & 45.05 / 99.09 & 42.25 / 98.61 & 40.64 / 98.31 & 38.93 / 97.88 & 36.34 / 97.15 & 40.63 / 98.21 \\
    \cellcolor{cyan!10}\methodnamenospace\textsubscript{ProCT}
    & \cellcolor{cyan!10}\textbf{45.53} / \textbf{99.27} & \cellcolor{cyan!10}\textbf{42.58} / \textbf{98.85} & \cellcolor{cyan!10}\textbf{40.98} / \textbf{98.57} & \cellcolor{cyan!10}\textbf{39.19} / \textbf{98.16} & \cellcolor{cyan!10}\textbf{36.65} / \textbf{97.52} & \cellcolor{cyan!10}\textbf{40.98} / \textbf{98.47} \\
    \shline
    \end{tabular}}
    \vspace{-6pt}
\end{table*}

\subsubsection{Implementation details}
\methodname is implemented in PyTorch and directly applied to both image-domain and dual-domain backbones, including FBPConvNet~\cite{fbpconvnet}, FredNet~\cite{gloredi}, ProCT~\cite{clma_proct}, and DuDoNet++~\cite{dudonet++}, to obtain the corresponding \methodname-enhanced variants.
Specifically, we insert \modelname{} in the image-domain network; for dual-domain networks, the sinogram-domain branch remains unchanged. 
For all backbones, we uniformly insert three \modelname{} modules into their image-domain encoders at three scales: $(H_0/2,\, W_0/2)$, $(H_0/4,\, W_0/4)$, and $(H_0/8,\, W_0/8)$.
For optimization, we use the Adam optimizer~\cite{adam} with $(\beta_1,\beta_2)=(0.5,0.999)$ to train the model for 90 epochs. The learning rate starts from $4\times 10^{-4}$ and is halved every 30 epochs. The model is trained on eight NVIDIA 4090 GPUs with a total batch size of 16.

\subsubsection{Evaluation metrics}
We use peak signal-to-noise ratio (PSNR) and structural similarity (SSIM)~\cite{ssim} as quantitative metrics. We report the average results of two commonly used CT windows: the full-range window [-1024, 3072] HU, and soft tissue window [-200, 300] HU.
To better showcase the effect of \methodname compared to the backbones, we calculate the normalized performance gain as follows:
\begin{align}
\Delta \% = \left( \frac{v_i - v_i^{\text{baseline}}}{v^{\text{ref}} - v_i^{\text{input}}} \right) \times 100,
\end{align}
where $v_i$ and $v_i^{\text{baseline}}$ represent the performance of \methodname and the corresponding baseline model, respectively. $v^{\text{ref}}$ denotes a reference upper value used only for normalization (50 dB for PSNR and 100$\%$ for SSIM), while $v_i^{\text{input}}$ is the value of the metal-affected input images as the lower bound.
Note that PSNR does not have a fixed theoretical maximum and can exceed 50 dB when the reconstruction error is very small. 
For clinical datasets without paired ground truth, we additionally report the Fréchet Inception Distance (FID)~\cite{fid} between the restored images and the artifact-free reference set.

\subsubsection{Comparison methods}
We compare our model with the following state-of-the-art methods: LI~\cite{LI}, NMAR~\cite{NMAR}, FBPConvNet~\cite{fbpconvnet}, FredNet~\cite{gloredi}, DuDoNet++~\cite{dudonet++}, ACDNet~\cite{hongacdnet}, QuadNet~\cite{quadnet}, RISE-MAR~\cite{risemar} and ProCT~\cite{clma_proct}. 
Among them, LI and NMAR are the widely used baselines in MAR, serving as classic MAR baselines. 
FBPConvNet and FredNet are image post-processing methods using a U-Net and Fourier network, respectively. ACDNet is an image-domain network utilizing a convolutional dictionary network to encode priors related to metal artifacts. 
RISE-MAR is a state-of-the-art semi-supervised framework that takes radiologists' feedback for enhanced performance. 
ProCT is a state-of-the-art image-domain network for CT reconstruction with an attention mechanism.  
DuDoNet++ and QuadNet are dual-domain methods, each employing separate networks in the sinogram and image domains to enhance the reconstruction quality.
All models are trained and tested on the same dataset for fair comparison. We utilized code from CNNMAR for LI and NMAR~\cite{cnnmar}, while employing ACDNet, FredNet, QuadNet, RISE-MAR, and ProCT according to the official code. For FBPConvNet and DuDoNet++, we carefully reproduced them following their original papers. 
To avoid domain gaps, all models are trained and tested on the same dataset under identical settings for fair comparison.

\begin{figure*}[!htb]
    \centering
    \includegraphics[width=1\linewidth]{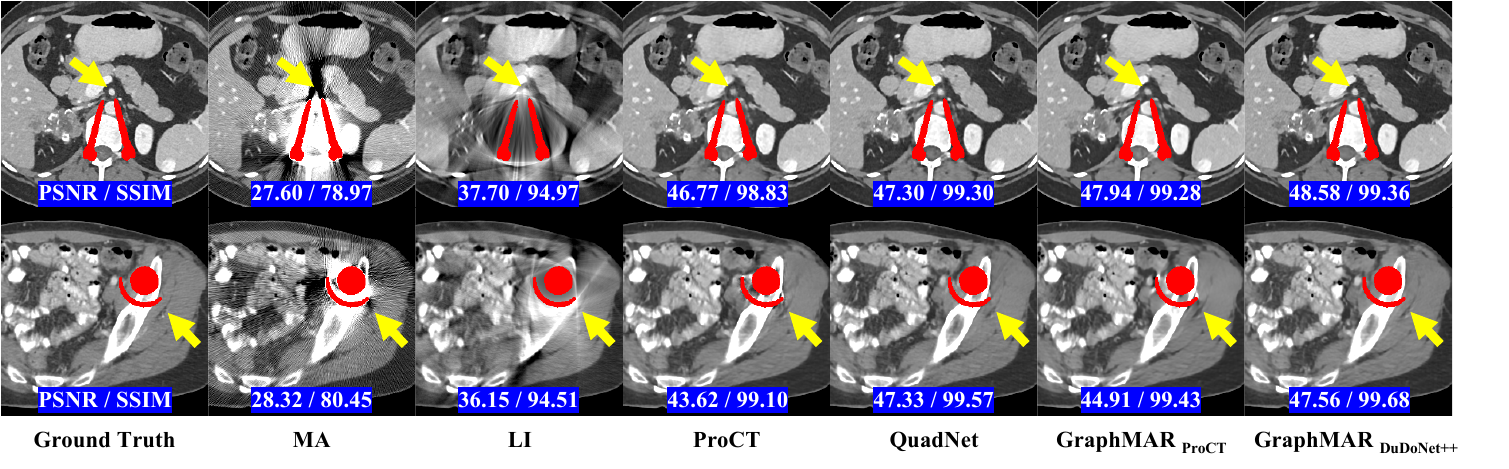}
    \vspace{-10pt}
    \caption{Qualitative comparison of different methods on the DDMAR dataset. Methods that are competitive in Table~\ref{tab:deeple} are selected for visual comparison.}
    \label{fig:compare_ddmar}
\end{figure*}

\begin{figure*}[!htb]
    \centering
    \includegraphics[width=1\linewidth]{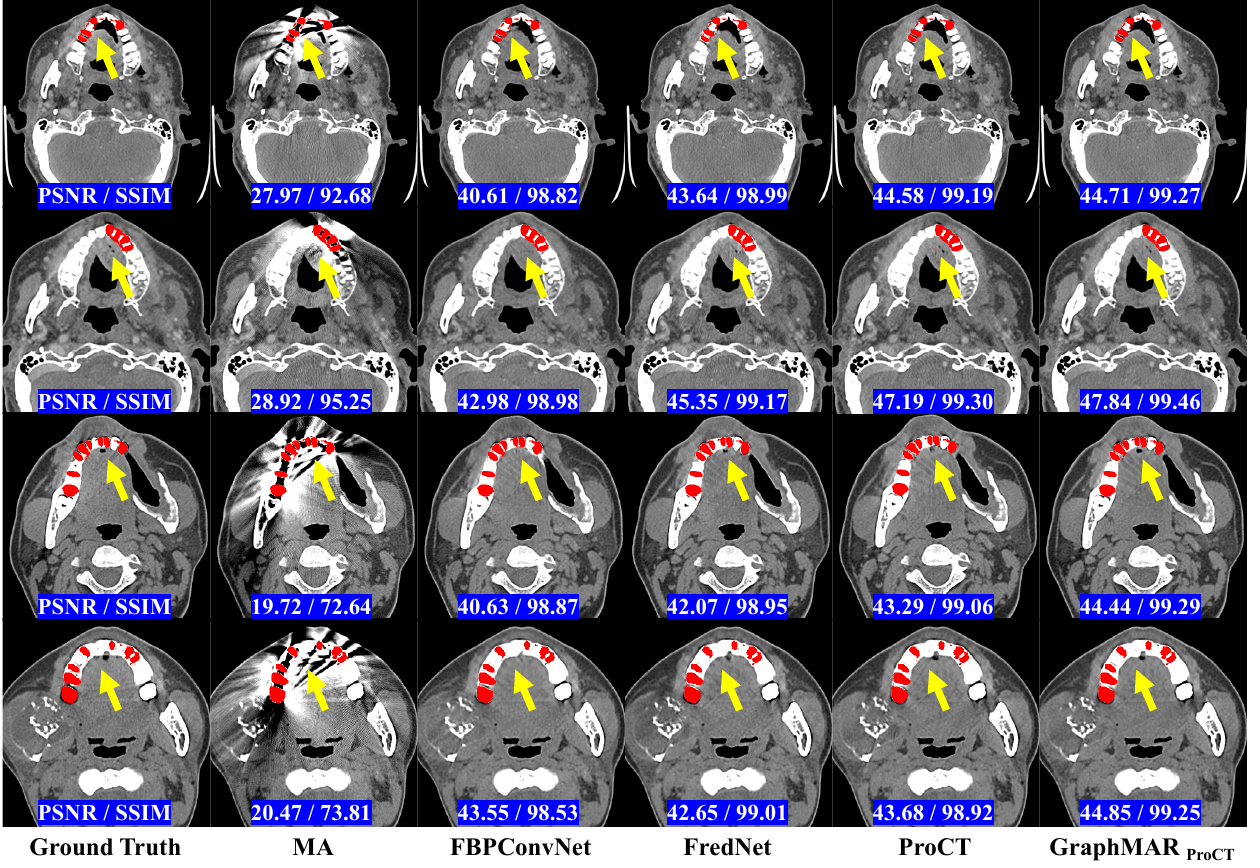}
    \vspace{-10pt}
    \caption{Qualitative comparison of different methods on the simulated dental dataset. Methods that are competitive in Table~\ref{tab:teeth} are selected for visual comparison.}
    \label{fig:compare_dental}
\end{figure*}

\begin{figure*}[t]
    \centering
    \includegraphics[width=1.0\linewidth]{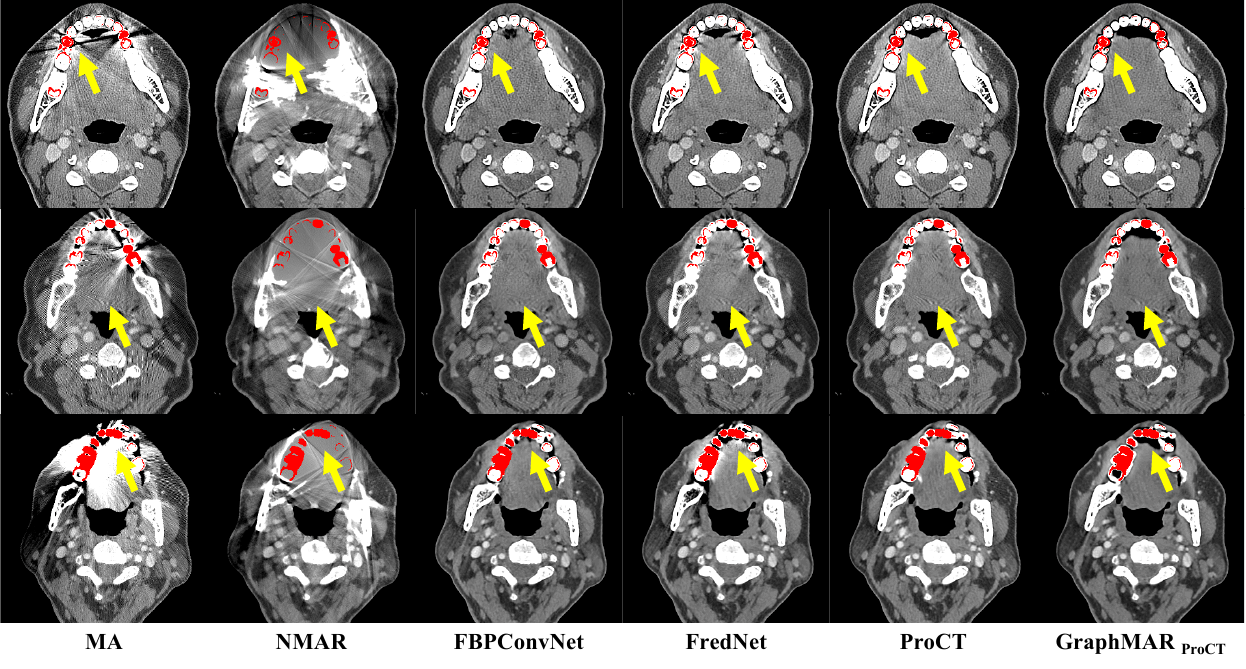}
    \caption{Qualitative comparison of different methods on the real-world dental CT with severe artifacts. Methods that are competitive in Table~\ref{tab:teeth} are selected for visual comparison.}
    \label{fig:compare_real}
\end{figure*}

\subsection{Results on Simulated Datasets}
\label{sec:result_simu}

\subsubsection{Quantitative results on the DDMAR dataset}
Table~\ref{tab:deeple} presents the quantitative results on the DDMAR dataset.
First, deep learning methods are generally superior to conventional interpolation-based methods, and dual-domain methods outperform single-domain ones owing to the additional sinogram information.
By integrating \methodname into both single-domain and dual-domain networks, we observe consistent improvements across all five metal-size groups.
Within the single-domain group, \methodnamenospace\textsubscript{ProCT} achieves the best average result, enhancing the backbone by over 1~dB in PSNR, which demonstrates the effectiveness of \methodname.
Within the dual-domain group, \methodnamenospace\textsubscript{DuDoNet++} also improves DuDoNet++ across different metal sizes, indicating that \methodname provides a complementary perspective from the graph beyond the conventional sinogram and image perspective.

\subsubsection{Quantitative results on the Dental dataset}
Table~\ref{tab:teeth} presents the results on the simulated Dental dataset.
Since raw sinograms are not available in this setting, which better reflects clinical scenarios where raw acquisition data cannot be obtained, sinogram-domain inputs are obtained via forward projection and deviate from raw measurements. We therefore focus on image-domain networks only and exclude methods that take sinogram priors (\eg, LI) as input.
First, while LI and NMAR improve the average outcomes, LI reduces the results for the smallest metal sizes compared to the input. This suggests that image-derived sinograms may introduce strong secondary artifacts when raw data is unavailable~\cite{quadnet}.
Second, from FBPConvNet, FredNet, to ProCT, which employ CNN, Fourier network, and transformer architectures respectively, we observe a notable performance progression, indicating that a larger receptive field and greater model capacity effectively improve MAR performance. 
Taking ProCT as the backbone, \methodnamenospace\textsubscript{ProCT} consistently improves the average performance, demonstrating the effectiveness of \methodname on the more challenging dental dataset.

\begin{table}[t]
    \centering
    \caption{FID comparison on the real-world Dental dataset. The lower the better. }
    \label{tab:fid_real}
    \begin{tabular}{l|c}
    \shline
    Method & FID $\downarrow$ \\
    \hline
    MA & 145.05 \\
    LI~\cite{LI} & 173.41 \\
    NMAR~\cite{NMAR} & 142.58 \\
    FBPConvNet~\cite{fbpconvnet} & 120.31 \\
    FredNet~\cite{gloredi} & 113.83 \\
    RISE-MAR~\cite{risemar} & 113.05 \\
    ProCT~\cite{clma_proct} & 115.83 \\
    \cellcolor{cyan!10}\methodnamenospace\textsubscript{ProCT} & \cellcolor{cyan!10}\textbf{108.73} \\
    \shline
    \end{tabular}
\end{table}

\subsubsection{Qualitative results on simulated datasets}
We select competitive or representative methods from Tables~\ref{tab:deeple} and~\ref{tab:teeth} for visual comparison.

Fig.~\ref{fig:compare_ddmar} shows visual results on the DDMAR dataset.
Conventional methods such as LI reduce artifacts but introduce secondary artifacts and blurry outputs.
Recent deep learning methods produce satisfactory results, yet still exhibit inaccuracies near the metal implants.
Specifically, in this ideal scenario where the original sinogram and precise CT geometry are available for reconstruction, dual-domain methods such as QuadNet still exhibit strong performance, especially when the metal implants are large.
By comparing \methodnamenospace\textsubscript{ProCT} with the backbone ProCT, we find that \methodnamenospace\textsubscript{ProCT} better restores fine details near the metal implants, such as vessels and soft tissues.
In addition, \methodnamenospace\textsubscript{DuDoNet++} achieves the best results, demonstrating that \methodname is effective for both image-domain and dual-domain methods.

Fig.~\ref{fig:compare_dental} shows results on the simulated Dental dataset, which is more challenging due to complex metal implants and stronger artifacts.
As ideal sinograms are not available in this setting, we exclude methods that rely on sinogram data or sinogram-based priors, such as LI and NMAR.
We find that all compared image-domain models reduce artifacts to some extent. Specifically, FBPConvNet and FredNet tend to over-smooth the details, producing inaccurate soft-tissue boundaries. ProCT yields better results, yet residual artifacts remain in some cases. \methodnamenospace\textsubscript{ProCT} achieves the closest results to the ground truth across all four cases.

In general, \methodname achieves the best visual quality on both datasets. We further discuss in Section~\ref{sec:interpretability} how the artifact attention produced by \methodname can be used for image post-processing to further enhance clinical reliability and reduce potential secondary artifacts and hallucinations.

\begin{table*}[t]
    \caption{Quantitative comparison in the form of [PSNR (dB) / SSIM (\%)] between backbone baselines and \methodname variants on the Dental dataset. Rows marked $\Delta \%$ report the normalized performance gain over the corresponding backbone, using 50 dB for PSNR and 100\% for SSIM as reference upper values. The best results are marked in \textbf{bold}.}
    \label{tab:backbone_ours_pergroup}
    \centering
    {%
    \setlength{\tabcolsep}{1.5pt}
    \begin{tabular}{l ccccc c}
    \shline
    Method & Small Metals & \multicolumn{3}{c}{$\longrightarrowtext{3cm}$} & Large Metals & Average \\
    \hline
    FBPConvNet
    & 40.54 / 98.85 & 38.56 / 98.37 & 37.38 / 98.07 & 36.10 / 97.63 & 34.16 / 96.88 & 37.35 / 97.96 \\
    \cellcolor{cyan!10}\methodnamenospace\textsubscript{FBPConvNet}
    &\cellcolor{cyan!10} 43.89 / 99.13 &\cellcolor{cyan!10} 41.09 / 98.66 &\cellcolor{cyan!10} 39.35 / 98.33 &\cellcolor{cyan!10} 37.60 / 97.86 &\cellcolor{cyan!10} 35.27 / 97.08 &\cellcolor{cyan!10} 39.44 / 98.21 \\
    $\Delta \%$
    & \textcolor{blue}{+20.6\%} / \textcolor{blue}{+10.3\%} & \textcolor{blue}{+12.0\%} / \textcolor{blue}{+5.7\%} & \textcolor{blue}{+8.5\%} / \textcolor{blue}{+3.9\%} & \textcolor{blue}{+5.8\%} / \textcolor{blue}{+2.5\%} & \textcolor{blue}{+3.8\%} / \textcolor{blue}{+1.5\%} & \textcolor{blue}{+9.1\%} / \textcolor{blue}{+3.4\%} \\
    \cmidrule{1-7}
    FredNet
    & 44.26 / 99.03 & 41.41 / 98.53 & 39.85 / 98.23 & 38.04 / 97.77 & 35.50 / 97.03 & 39.81 / 98.12 \\
    \cellcolor{cyan!10}\methodnamenospace\textsubscript{FredNet}
    &\cellcolor{cyan!10} 44.69 / 99.21 &\cellcolor{cyan!10} 41.61 / 98.75 &\cellcolor{cyan!10} 39.97 / 98.44 &\cellcolor{cyan!10} 38.13 / 97.98 &\cellcolor{cyan!10} 35.56 / 97.23 &\cellcolor{cyan!10} 39.99 / 98.32 \\
    $\Delta \%$
    & \textcolor{blue}{+2.6\%} / \textcolor{blue}{+6.6\%} & \textcolor{blue}{+1.0\%} / \textcolor{blue}{+4.3\%} & \textcolor{blue}{+0.5\%} / \textcolor{blue}{+3.1\%} & \textcolor{blue}{+0.3\%} / \textcolor{blue}{+2.3\%} & \textcolor{blue}{+0.2\%} / \textcolor{blue}{+1.5\%} & \textcolor{blue}{+0.8\%} / \textcolor{blue}{+2.7\%} \\
    \cmidrule{1-7}
    ProCT
    & 45.05 / 99.09 & 42.25 / 98.61 & 40.64 / 98.31 & 38.93 / 97.88 & 36.34 / 97.15 & 40.63 / 98.21 \\
    \cellcolor{cyan!10}\methodnamenospace\textsubscript{ProCT}
    & \cellcolor{cyan!10}45.53 / 99.27 & \cellcolor{cyan!10}42.58 / 98.85 &\cellcolor{cyan!10} 40.98 / 98.57 &\cellcolor{cyan!10} 39.19 / 98.16 &\cellcolor{cyan!10} 36.65 / 97.52 &\cellcolor{cyan!10} 40.98 / 98.47 \\
    $\Delta \%$
    & \textcolor{blue}{+2.9\%} / \textcolor{blue}{+6.6\%} & \textcolor{blue}{+1.6\%} / \textcolor{blue}{+4.7\%} & \textcolor{blue}{+1.5\%} / \textcolor{blue}{+3.9\%} & \textcolor{blue}{+1.0\%} / \textcolor{blue}{+3.0\%} & \textcolor{blue}{+1.1\%} / \textcolor{blue}{+2.8\%} & \textcolor{blue}{+1.5\%} / \textcolor{blue}{+3.5\%} \\
    \cmidrule{1-7}
    DuDoNet++
    & 42.67 / 98.90 & 40.56 / 98.45 & 39.12 / 98.12 & 37.52 / 97.66 & 35.36 / 96.84 & 39.04 / 97.99 \\
    \cellcolor{cyan!10}\methodnamenospace\textsubscript{DuDoNet++}
    & \cellcolor{cyan!10}\textbf{46.99} / \textbf{99.44} & \cellcolor{cyan!10}\textbf{44.53} / \textbf{99.12} & \cellcolor{cyan!10}\textbf{43.08} / \textbf{98.91} & \cellcolor{cyan!10}\textbf{41.49} / \textbf{98.61} & \cellcolor{cyan!10}\textbf{39.00} / \textbf{98.08} & \cellcolor{cyan!10}\textbf{43.01} / \textbf{98.83} \\
    $\Delta \%$
    & \textcolor{blue}{+26.5\%} / \textcolor{blue}{+19.9\%} & \textcolor{blue}{+18.9\%} / \textcolor{blue}{+13.1\%} & \textcolor{blue}{+17.0\%} / \textcolor{blue}{+11.7\%} & \textcolor{blue}{+15.4\%} / \textcolor{blue}{+10.3\%} & \textcolor{blue}{+12.6\%} / \textcolor{blue}{+9.5\%} & \textcolor{blue}{+17.2\%} / \textcolor{blue}{+11.4\%} \\
    \shline
    \end{tabular}}
    \vspace{-4pt}
\end{table*}

\subsection{Generalization on Real-World Datasets}
\label{sec:result_real}
We use the checkpoint of each model trained on the simulated Dental dataset and directly generalize to the real-world dental CT with metal implants.  
Since paired artifact-free references are unavailable, we use the artifact-free images without metal implants as a reference set and report FID for quantitative results.

Table~\ref{tab:fid_real} shows the FID of each method. First, we note that LI achieves a higher (worse) FID than the uncorrected MA input, confirming that sinogram-domain methods relying on image-derived projections can introduce severe secondary artifacts and degrade image quality. For deep learning methods, we find that \methodname achieves the best result and enhances the performance of the backbone ProCT.

Fig.~\ref{fig:compare_real} shows the visual results on real-world dental CT.
FBPConvNet and FredNet produce visually implausible results, such as unrealistic tongue shapes and residual streak artifacts.
ProCT better reduces global artifacts but fails to fully remove artifacts around the metal implants.
\methodname is more effective at reducing artifacts overall. In addition, benefiting from the extra perspective provided by the geometric graph, \methodname produces more reasonable anatomical fine structures in severely affected regions near the metal and clearly outperforms the backbone ProCT.
The results demonstrate the effectiveness of \methodname in clinical applications.

\subsection{Ablation Studies}
\label{sec:ablation}
In this section, we study the detailed design of \methodname.
Unless noted otherwise, all ablation experiments are conducted with \methodnamenospace\textsubscript{FBPConvNet} on the Dental dataset, as it is the most widely adopted backbone model.

\subsubsection{Ablation on different backbone models}
\label{sec:ablation_backbone}
We study whether \methodname can adapt to different backbone models.
Table~\ref{tab:backbone_ours_pergroup} reports the results of four backbone models, covering both image-domain and dual-domain networks, with and without \methodname.
First, \methodname consistently improves all tested backbones, including both image-domain and dual-domain methods.
Second, the normalized gain $\Delta\%$ is largest for FBPConvNet and DuDoNet++, whose convolutional encoders have limited receptive fields and therefore benefit most from the global geometric guidance.
Third, we find \methodname is also effective on models with a global receptive field, such as FredNet and ProCT using Fourier convolution or transformer, respectively.
The consistent performance gain across diverse architectures indicates that \methodname provides complementary information from a graph perspective that is essential for MAR yet overlooked in previous studies.

\begin{table}[t]
    \centering
    \small
    \caption{Ablation of each component of \methodname on the Dental dataset.}
    \begin{tabular}{c|ccc|cc}
    \shline
    Config & Graph router & MoE & $\mathcal{L}_{\mathrm{graph}}$ & PSNR & SSIM \\
    \hline
    \textbf{a)} & \checkmark & \checkmark & \checkmark & \textbf{39.44} & \textbf{98.21} \\
    \textbf{b)} & \checkmark & & \checkmark & 38.69 & 98.14 \\
    \textbf{c)} & \checkmark & \checkmark & & 38.26 & 98.07 \\
    \textbf{d)} & & \checkmark & \checkmark & 37.94 & 98.04 \\
    \textbf{e)} & &  & \checkmark & 37.82 & 98.01 \\
    \textbf{f)} & & & & 37.35 & 97.96 \\
    \shline
    \end{tabular}
    \label{tab:ablation_overall}
\end{table}

\begin{table}[t]
    \centering
    \small
    \caption{Ablation on polar-coordinate artifact graph.}
    \begin{tabular}{c|cc|cc}
    \shline
    Config & Angular & Radial & PSNR & SSIM \\
    \hline
    \textbf{a)} & \checkmark & \checkmark & \textbf{39.44} & \textbf{98.21} \\
    \textbf{b)} & & \checkmark & 38.59 & 98.11 \\
    \textbf{c)} & \checkmark & & 38.51 & 98.05 \\
    \textbf{d)} & & & 37.73 & 97.88 \\
    \shline
    \end{tabular}
    \label{tab:ablation_graph}
\end{table}

\begin{table}[t]
    \centering
    \small
    \caption{Ablation on the number $K$ of experts.}    \begin{tabular}{c|ccccc}
    \shline
    $K$ & 1 & 2 & 3 & 4 & 5 \\
    \hline
    PSNR & 38.69 & 38.92 & \textbf{39.44} & 38.81 & 38.79 \\
    SSIM & 98.14 & 98.17 & \textbf{98.21} & 98.05 & 98.04 \\
    \shline
    \end{tabular}
    \label{tab:ablation_experts}
\end{table}

\begin{figure*}[t]
    \centering
    \includegraphics[width=.8\linewidth]{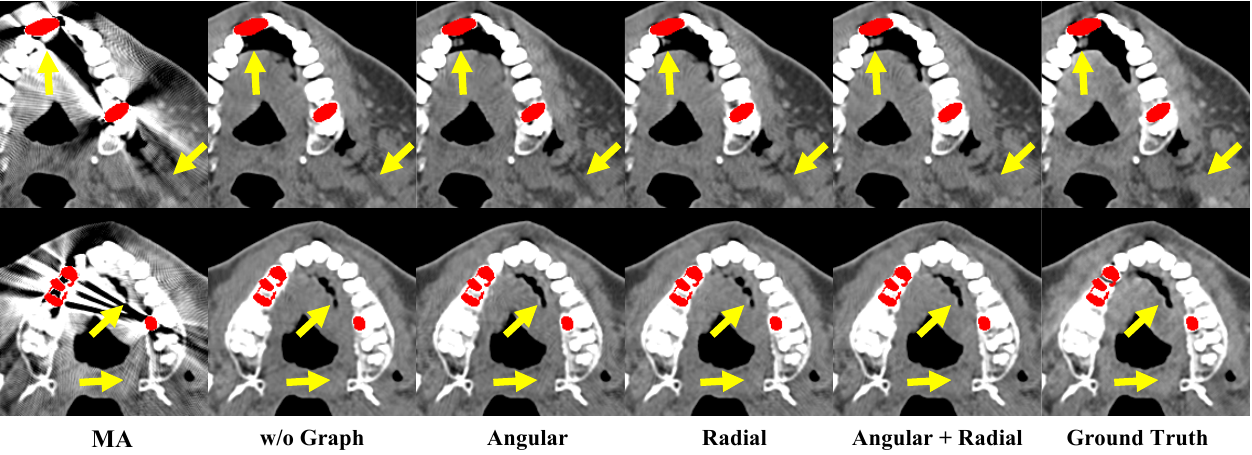}
    \caption{Qualitative comparison of different configurations of \methodname in Table~\ref{tab:ablation_graph}. Zoom in for a better view.}
    \label{fig:ablation_graph}
\end{figure*}

\begin{table}[t]
    \centering
    \small
    \caption{Ablation on the geometric alignment loss $\mathcal{L}_{\mathrm{graph}}$.}
    \begin{tabular}{l|cc}
    \shline
    Alignment loss & PSNR & SSIM \\
    \hline
    MSE & \textbf{39.44} & \textbf{98.21} \\
    KL & 38.49 & 98.10 \\
    w/o $\mathcal{L}_{\mathrm{graph}}$ & 38.26 & 98.07 \\
    \shline
    \end{tabular}
    \label{tab:ablation_loss}
\end{table}

\begin{table}[t]
    \centering
    \small
    \caption{Ablation on plugin insertion position.}
    \begin{tabular}{c|ccc|cc}
    \shline
    Config & $H/2$ & $H/4$ & $H/8$ & PSNR & SSIM \\
    \hline
    \textbf{a)} & \checkmark & \checkmark & \checkmark & \textbf{39.44} & \textbf{98.21} \\
    \textbf{b)} & & \checkmark & \checkmark & 39.12 & 98.16 \\
    \textbf{c)} & & & \checkmark & 38.91 & 98.10 \\
    \textbf{d)} & & \checkmark & & 38.40 & 98.05 \\
    \shline
    \end{tabular}
    \label{tab:ablation_position}
\end{table}

\subsubsection{Ablation of each component in \methodname}
\label{sec:ablation_overall}
Table~\ref{tab:ablation_overall} presents the results of removing each component in \methodname.
For configuration (b), we use a single expert to substitute the mixture-of-experts, and for configurations (d) and (e) without the graph router, we use a convolution layer to replace it.
First, by comparing (b) and (c) to (a), we find that both the MoE module and geometric alignment are important for \methodname; moreover, the graph router itself significantly enhances the result over the plain backbone (f).
Second, by comparing (d) and (e) to (a), we notice that even a simple convolution with geometric alignment is effective for MAR, indicating that the geometric alignment loss serves as an important supervisory signal that helps the model focus on artifact-affected regions.
In general, our full model achieves the best result and significantly enhances the performance of the backbone.

\begin{table}[t]
    \centering
    \footnotesize
    \caption{Efficiency of each method, including the number of parameters, FLOPS, and inference time. The test is conducted on a single RTX 4090 GPU using $1000$ images with a batch size of 1, each at a resolution of $512 \times 512$.}
    \begin{tabular}{l|rrr}
    \shline
    Method & Param. (M) & FLOPS (G) & Infer. (ms) \\
    \hline
    FBPConvNet~\cite{fbpconvnet} & 17.27 & 160.13 & 12.34 \\
    \methodnamenospace\textsubscript{FBPConvNet} & 20.03 & 224.06 & 25.77 \\
    \hline
    DuDoNet++~\cite{dudonet++} & 27.45 & 298.99 & 26.22 \\
    \methodnamenospace\textsubscript{DuDoNet++} & 30.21 & 362.77 & 40.03 \\
    \hline
    FredNet~\cite{gloredi} & 9.69 & 227.62 & 24.67 \\
    \methodnamenospace\textsubscript{FredNet} & 11.53 & 252.37 & 48.73 \\
    \hline
    ProCT~\cite{clma_proct} & 2.48 & 98.00 & 51.45 \\
    \methodnamenospace\textsubscript{ProCT} & 3.11 & 118.69 & 89.45 \\
    \shline
    \end{tabular}
    \label{tab:efficiency}
\end{table}

\begin{figure}[t]
    \centering
    \includegraphics[width=1\linewidth]{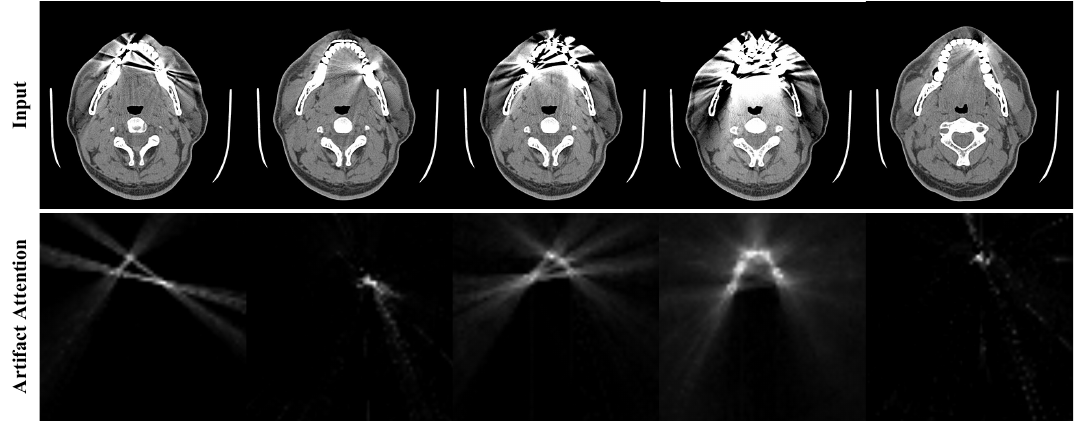}
    \caption{The output artifact attention map $\mat{G}_A$ produced by \methodname on the Dental dataset. Brighter regions indicate higher artifact severity. Zoom in for a better view.}
    \label{fig:artifact_attention}
\end{figure}

\begin{figure*}[!t]
    \centering
    \includegraphics[width=1\linewidth]{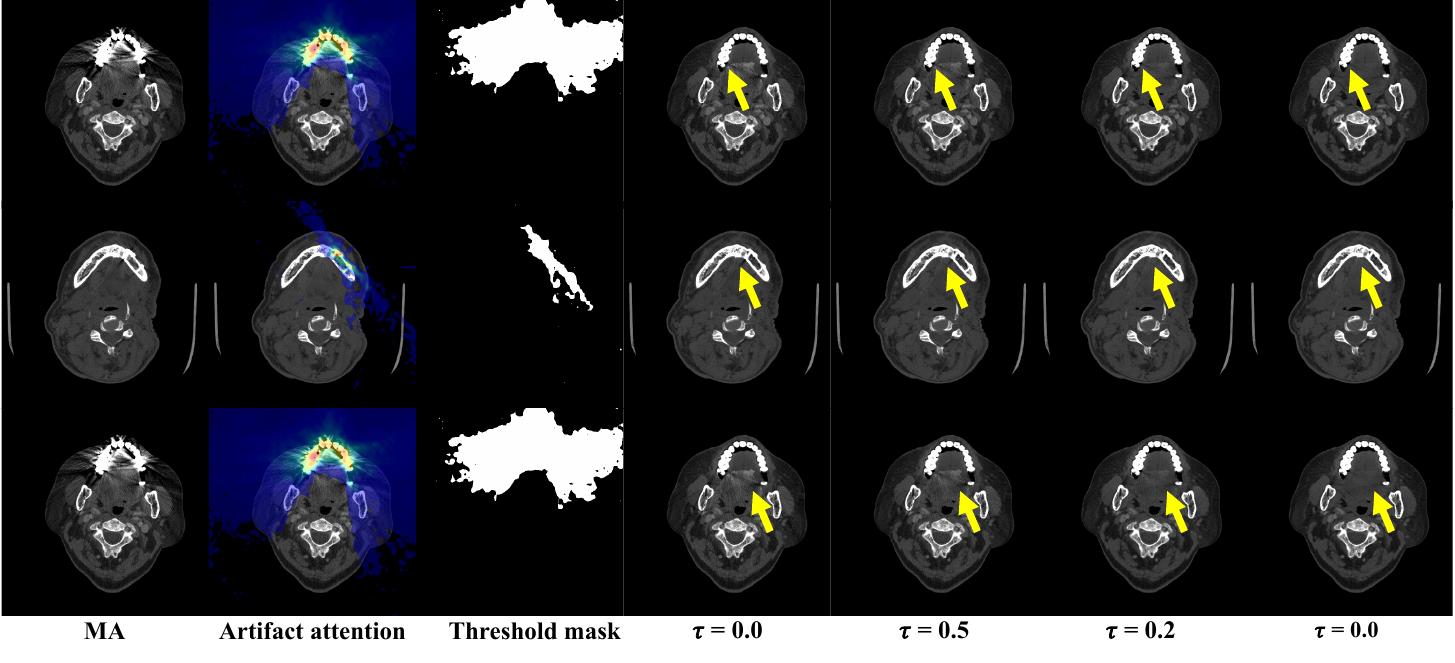}
    \caption{Clinical application of artifact attention. The artifact attention $\mat{G}_A$ localizes regions where the model focuses on. By thresholding $\mat{G}_A$, a binary mask $\mat{M}_A$ separates severely affected regions from lightly affected ones. A user-defined blending parameter $\tau$ then controls the fusion strength in the remaining regions, enabling selective fusion between the MAR output and the original input to improve clinical reliability.}
    \label{fig:clincal_fuse}
\end{figure*}

\subsubsection{Effect of the polar-coordinate artifact graph}
\label{sec:ablation_graph}
The polar-coordinate artifact graph in \modelname connects patch nodes via two types of polar edges: angular edges that link nodes sharing similar streak directions, and radial edges that link nodes at similar distances to the metal.
Table~\ref{tab:ablation_graph} compares different edge constructions.
We find that removing either the angular or the radial edges reduces performance, indicating that the two edge types capture complementary aspects of artifact geometry.
Note that removing both edge types degenerates the graph into isolated nodes where no message passing occurs between patches, equivalent to per-patch independent processing, thus yielding the worst result.

Fig.~\ref{fig:ablation_graph} illustrates the visual results of different configurations in Table~\ref{tab:ablation_graph}. We find that angular edges significantly enhance the details between implants and suppress streak artifacts, as indicated by the upper arrows in both samples. Radial edges, on the other hand, restore the dark regions by borrowing information from pixels at similar radial distances, as indicated by the lower arrows in the first row. 
These results demonstrate the effectiveness of the proposed artifact graph construction, and further indicate that graph-based geometric knowledge provides essential structural guidance for MAR that is difficult to obtain from conventional networks.

\subsubsection{Effect of the number of experts.}
Table~\ref{tab:ablation_experts} shows the results with different numbers of experts.
Using too few experts limits the model's capacity to capture diverse degradation patterns, while too many experts may introduce redundant branches.  
The results show that $K=3$ achieves the best performance.

\subsubsection{Effect of the geometric alignment loss}
\label{sec:ablation_loss}
The geometric alignment loss $\mathcal{L}_{\mathrm{graph}}$ supervises the learned routing map to align with the geometric prior derived from the metal mask.
Table~\ref{tab:ablation_loss} compares different supervision strategies.
In general, we find pixel-level alignment is more suitable for supervising the spatially structured routing map.

\subsubsection{Effect of plugin position}
\label{sec:ablation_position}
We insert \modelname at the image-domain encoder of the backbone. Here $H/k$ denotes the feature map at resolution $(H_0/k,\, W_0/k)$.
Table~\ref{tab:ablation_position} compares different insertion configurations.
First, a single plugin at $H/8$ outperforms one at $H/4$, suggesting that deeper, lower-resolution features benefit more from geometric guidance. This is likely because artifact traces become attenuated and smoothed after repeated convolution in deeper layers, while \methodname further highlights these weakened artifact patterns through explicit geometric routing. 
Second, progressively adding higher-resolution scales ($H/4$, then $H/2$) further improves the result, and the three-scale setting achieves the best performance.

\subsubsection{Efficiency}
Table~\ref{tab:efficiency} reports the computational overhead of \methodname on different backbone models.
First, we find that the increased parameters and FLOPs are modest, and the overall cost remains much lower than adding a separate sinogram-domain network.
Second, we notice that \methodname leads to slower inference speed, mainly due to the graph construction and other conventional image processing operations, which are not parallelizable on the GPU.
However, the additional latency is less than 100 ms and has a negligible impact on practical deployment.
In general, \methodname is lightweight and can effectively enhance the backbone model without relying on sinogram data or larger architectures, making it practical for clinical use.

\subsection{Interpretability and Clinical Reliability}
\label{sec:interpretability}

Fig.~\ref{fig:artifact_attention} shows the artifact attention $\mat{G}_A$ produced by \modelname, which highlights artifact-affected regions and aligns well with human observation. This provides an interpretable indication of where the model focuses during MAR, improving the reliability of MAR results. 

Furthermore, $\mat{G}_A$ enables selective restoration that conventional end-to-end MAR methods cannot offer: existing methods process the entire image uniformly, leaving users unaware of whether secondary artifacts or hallucinations have been introduced. Since \methodname explicitly localizes artifact regions, MAR can be applied only where needed.

As shown in Fig.~\ref{fig:clincal_fuse}, we obtain a binary threshold mask $\mat{M}_A$ by thresholding $\mat{G}_A$ to separate pixels into severely affected and less affected regions. The fused output combines the MAR result $\hat{Y}$ and the original input $X$ via:
$\hat{Y}_{\mathrm{fuse}} = \mat{M}_A \odot \hat{Y} + (\mat{1} - \mat{M}_A) \odot \big( \tau \hat{Y} + (1-\tau) X \big)$,
where $\odot$ denotes element-wise multiplication, and $\tau \in [0,1]$ is a user-adjustable parameter to control the restoration strength in the less affected regions.
By applying MAR only where artifacts are severe and allowing adjustable blending elsewhere, this fusion reduces the risk of hallucination and secondary artifacts, thereby improving clinical reliability. Moreover, by adjusting $\tau$, users can clearly observe how artifacts are progressively reduced, which enhances transparency and strengthens confidence in the MAR result.

In summary, \methodname not only improves MAR performance but also offers a new perspective and application through explicit artifact localization. More advanced fusion strategies can be developed based on \methodname in the future, enabling clinicians to flexibly control the MAR results according to their diagnostic needs.

\section{Discussion}

Deep learning has driven remarkable progress in metal artifact reduction for CT imaging. Yet, existing approaches still implicitly couple artifact identification and reduction, which not only makes the model prone to over-smoothing in artifact-free regions, but also leaves the MAR process uninterpretable and restricts the flexibility for selective restoration in clinical use.

To address these limitations, we introduce explicit artifact geometry into image-domain networks. By constructing a geometric density graph and a polar-coordinate artifact graph, \methodname decouples artifact identification from reduction. This enables spatially adaptive restoration via graph-guided routing, which processes regions of varying artifact severity in a divide-and-conquer manner.
\methodname benefits MAR in two aspects.
First, \methodname consistently improves backbone performance while operating entirely in the image-domain feature space. It does not require raw sinogram inputs, making it practical for clinical scenarios where raw data is unavailable.
Second, \methodname explicitly highlights artifact regions through the artifact attention, providing interpretable artifact identification absent in existing MAR methods.

We also acknowledge some limitations.
First, the geometric graph relies on inter-implant geometry and becomes less informative when only a single implant is present, as no inter-implant lines can be formed. In such cases, \modelname only utilizes radial and angular geometry around the implant to model the artifact. 
Second, although \methodname is lightweight, the artifact graph construction and graph-guided routing introduce additional computation. 
Third, this study focuses on validating the effectiveness of geometry-aware graph learning as a general framework; future work could refine the geometric graph for more complex clinical cases and explore advanced graph architectures to further enhance the performance.

\section{Conclusion}
In this paper, we introduce \methodname, a geometry-aware graph learning framework that enhances existing MAR networks with explicit artifact identification and spatially adaptive MAR.
Extensive experiments show that \methodname consistently improves diverse backbone models. As the first work to introduce graph-based modeling for MAR, \methodname enables explicit artifact identification in the image domain. We hope \methodname inspires further exploration of graph-based approaches for MAR and other tasks in the field of medical image analysis.




\bibliographystyle{IEEEtran}
\bibliography{reference}

\end{document}